\documentclass[12pt]{article}

\usepackage{amssymb,amsmath}
\usepackage{epsfig}
\usepackage{epsf}
\usepackage{float}
\usepackage{a4wide}
\usepackage{graphpap}
\usepackage{setspace}

\usepackage{latexsym}
\usepackage{graphicx}
\usepackage{array}
\usepackage{hhline}
\usepackage{fullpage}          % para aprovechar mejor la hoja
\usepackage{epsfig}
\newtheorem{thm}{Theorem}[section]

\usepackage[ruled,titlenotnumbered]{algorithm2e}

\newcommand{\be}{\begin{equation}}
\newcommand{\ee}{\end{equation}}
\newcommand{\ben}{\begin{eqnarray}}
\newcommand{\een}{\end{eqnarray}}

\usepackage{mathrsfs}
\usepackage{amssymb}
\usepackage{verbatim}
\usepackage{amsmath}

\usepackage{subfigure}

\usepackage{caption}
\usepackage{multirow}

\usepackage{graphicx}
\usepackage{subfigure}
\usepackage{amssymb,amsmath}
\usepackage{bm,bbm}
\usepackage{booktabs}
\usepackage{subfigure}
\usepackage{microtype}
\usepackage{color}

\title{Markovian models for one dimensional structure estimation on heavily noisy imagery.}

\author{Ana Georgina Flesia Javier Gimenez and Elena Rufeil Fiori\\FaMAF-UNC\\Ing. Medina Allende s/n, Ciudad Universitaria\\ CP 5000, C\'ordoba, Argentina. }

\begin{document}
\maketitle
\begin{abstract}

Radar (SAR) images often exhibit profound appearance variations
due to a variety of factors including clutter noise produced by the coherent nature of the illumination. Ultrasound images and infrared images have similar cluttered appearance, that make 1 dimensional structures, as edges and object boundaries difficult to locate. Structure information is usually  extracted in two steps: firstly, building and edge strength mask classifying pixels as edge points by hypothesis testing, and secondly computing on pixel wide connected edges. With constant false alarm rate (CFAR) edge strength detectors for speckle clutter, the image needs to be scanned by a sliding window composed of several differently oriented splitting sub-windows. The accuracy of edge location for these ratio detectors depends strongly on the orientation of the sub-windows.   In this work we propose to transform the edge strength detection problem into a binary segmentation problem  in the undecimated wavelet domain, solvable using parallel 1d Hidden Markov Models. For general dependency models, exact estimation of the state map becomes computationally complex, but in our model, exact MAP is feasible. The effectiveness of our approach is demonstrated on simulated noisy real-life natural images with available ground truth, while the strength of our output edge map is measured with Pratt's, Baddeley an Kappa proficiency measures. Finally, analysis and experiments  on three different types of SAR images, with different polarizations, resolutions and textures, illustrate that the proposed method can detect structure on SAR images effectively, providing a very good start point for active contour methods.

\end{abstract}

Keywords:Haar wavelet edge detection; Graphical models, SAR images; undecimated wavelet transform

%%%%%%%%%%%%%%%%%%%%%%
%%%%%%%%%%%%%%%%%%%%%%
\section{Introduction}
%%%%%%%%%%%%%%%%%%%%%%
%%%%%%%%%%%%%%%%%%%%%%
Satellite born Synthetic Aperture Radar (SAR) images  represent the complex reflectivity map of a scene, and unlike optical images, the interpretation is not consistent with common visual perception. Furthermore, the direct application of conventional image processing tools conceived from an optical point of view usually gives suboptimum results on SAR data. Among the most important differences between optical and SAR images are the appropriate noise models. In special, a complex SAR image may be represented as the convolution of the local complex reflectivity of the observed area with the impulse response of the SAR system. Hence, the SAR data pixels are roughly the low pass filtered version of the complex local scattering properties of the observed scene. A usual model for the complex image is obtained considering its real and imaginary parts as independent, zero mean and equally distributed random variables. Consequently, the intensity of a SAR image follows a negative exponential distribution, and can be further modeled as a product of the radar cross section and  a noise term called speckle, which is exponentially distributed with mean equal to one, Tello Alonso et al. (2011).
In the framework of SAR processing, it is useful to take the logarithm of the original signal in order to manage the multiplicative speckle; thus, after this transformation, the speckle is not only additive and signal independent, but also approximately Gaussian. Moreover, the logarithm operation is helpful in reducing the large dynamic range of SAR data.

A crucial point for SAR image automatic interpretation is the low level step of scene segmentation, i. e. the decomposition of the
image in a tessellation of uniform areas. Edge-based segmentation schemes aim to find out the transitions between  uniform areas, rather than directly identifying them. The related algorithms generally work in two stages: they firstly compute an edge strength map of the scene and finally
extract the local maxima of this map. The first stage is usually achieved using an  statistical edge detection filter. It consists in scanning the image with a sliding two-region window and evaluating for each position if there is a change between the two regions by hypothesis testing. The test are dependant of the true underlying conditional distributions that are emitted by the states (classes) of the segmentation field.

Our algorithm for edge strength mask detection in heavily noisy images, like log-amplitude satellite born Synthetic Aperture Radar (SAR) images, which is proposed in this paper, relies on the difference of statistical behavior along the wavelet scales of the Gaussian noise in front of the edges. The discontinuities are highlighted by the wavelet transform, and they tend to persist over scales. These two key properties, very well known  in the optical image processing literature, were described by Romberg et al. (2001) as:
\begin{enumerate}
\item Two Populations:  Smooth  signal/image  regions are represented  by  small  wavelet  coefficients, while edges,
ridges, and other singularities are represented by large coefficients.
\item Persistence:  Large  and  small  wavelet  coefficient values
cascade along the scale/space representation.
\end{enumerate}
Interpreting these properties statistically, Two Populations imply that wavelet coefficients have non-Gaussian
marginal statistics, with  a large peak at zero (many small coefficients) and heavy tails (a few large coefficients), while Persistence implies that wavelet coefficient values are statistically dependent along the scale/space representation. Nevertheless, the critical sub-sampling  in the Discrete Wavelet transform (DWT), which discards every second wavelet coefficient at each decomposition level,  results in wavelet coefficients that are highly dependent on their location in the sub-sampling lattice. Small shifts in the input waveform may cause large changes in the wavelet coefficients, large variations in the distribution of energy at different scales, and possibly large changes in reconstructed waveforms.

In Tello Alonso et al. (2011), the authors advocate for the use of the undecimated wavelet transform for detecting and enhancing edges on SAR images. They propose a deterministic multiscale thresholding scheme to produce an edge strength mask that single out edge candidates to apply snake algorithms in order to get one pixel wide continuous edges. This combination algorithms, in the radiometric domain, are popular in the literature of SAR edge detection. Radiometric edge strength masks are usually obtained by hypothesis testing, using the likelihood ratio filter, the robust rank order or the Wilcoxon test, or costume made maximum likelihood ratio tests (see Lim et al (2006), Germain et al (2001), Giron et al (2012) and references therein).

Our approach is more automatic than the one introduced by Tello Alonso et al (2011), since we estimate all parameters from the data using a Bayesian model in the undecimated wavelet domain. We model each string of pixel related wavelet coefficients as independent 1d Hidden Markov Model with Laplacian and Gaussian marginal distributions, related to  ``edge'' and ``no-edge'' states, respectively, as in Wang et al (2009). We derive our own MAP equations to compute the model parameters in order to predict intrascale and interscale edge strength maps. Then we combine these maps into one edge strength map that can be processed with standard Gradient based edge detection algorithms, watershed algorithms or active contour algorithms.
\begin{algorithm}
\caption{Bayesian edge detector}
\begin{itemize}
\item[-] Compute $W(m,n)$ undecimated wavelet transform, obtaining $T$ matrices of coefficients with the same size that the original image $I$ per directional band.
\item[-] Separate the Vertical and Horizontal Bands to model them independently.
\item[-] Within each band, model each chain of pixel related coefficient as a 1d Hidden Markov Chain with hidden states ``edge'' and ``no-edge''.
\item[-] Compute the model and evaluate the most likely state map for each chain.
\item[-] Combine the resulting maps with the intra-scale and inter-scale operators, producing an edge strength  map.
\item[-] Refine the edge strength map to produce  a one size pixel edge map.
\end{itemize}
\end{algorithm}

 In Sections 2 and 3 we introduce wavelet transforms and the Bayesian modeling of wavelet coefficients, leaving the proof of the theorems to Appendix A. In Section 4 we discuss processing of the obtained esge strenght mask, and develop some synthetic experiments, adding noise to good quality optical images with available ground truth. Edge quality is measured with three different supervised edge measures, Pratt's FOM, Baddely and Kappa. We also provide experiments with benchmark SAR images, that compare with other results from the literature.   The conclusions and prospects are left for for Section 5.
%%%%%%%%%%%%%%%%%%%%%%%%%%%%%%%%%%%%%
\section{Undecimated Haar wavelet transform}

The undecimated discrete wavelet transform (UDWT) has been independently discovered several times for different purposes and under different names, shift/translation invariant wavelet transform, stationary wavelet transform or redundant wavelet transform. The key point is that it is redundant, shift invariant, and gives an approximation to the continuous wavelet transform  denser than the approximation given by the discrete wavelet transform (DWT), Guo et al. (1995). The most cited implementations of this idea are Mallat's  \textit{\`a trous} algorithm
%that intercalate zero rows and columns in the filters before applying the same cascade algorithm as the DWT, and
and Beylkin's algorithm; for details, see Starck et al. (2007) and references therein.  % that emulates all possible shifts in the signal and average the results.

In our implementation, we use an algorithm based on the idea of no decimation.
%It applies the wavelet transform and omits both down sampling in the forward and up sampling in the inverse transform. More precisely, it applies the transform at each point of the image and saves the detail coefficients and uses the low-frequency coefficients for the next level.
The size of the coefficients array do not diminish from level to level, so by using all coefficients at each level, we get very
well allocated high-frequency information. In the UDWT algorithm, the number of pixels involved in computing a given coefficient grows slower and so the relation between the frequency and spatial information is more precise, Gyaourova et al. (2001).
\begin{figure}[h]
\includegraphics[width=15cm]{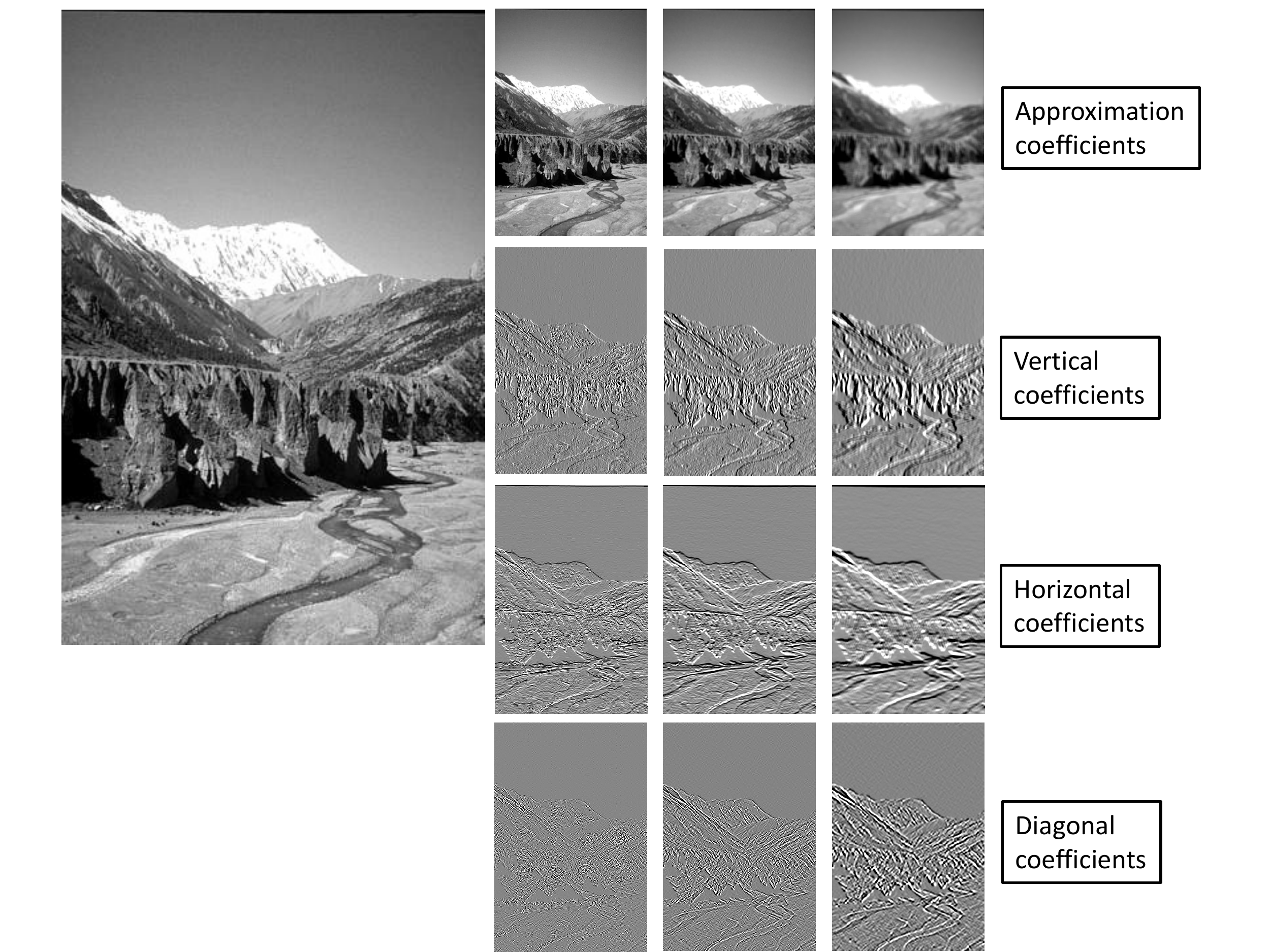}%\includegraphics[width=8cm]{Paper_StatCOMP_fig3.pdf}
\caption{ UDWT of ``mountain'' image with maximum decomposition  level equal to three. The undecimated wavelet transform allows to link the pixel with the coefficient, beside giving better edge localization than the DWT. }%and b) DWT decomposition, both
\label{mountain}
\end{figure}

The undecimated wavelet transform can be defined with all types of wavelet filters, but we use the Haar wavelet since its analyzing filter (the shortest possible) gives good localization and poor suppression of discontinuities.  Tello Alonso et al. (2011) consider Beylkin's stationary Haar wavelet transform for edge detection on SAR data. Thresholding Beylkin's algorithm produces very smooth results, which makes the images to look a little blurred. On the other hand, it makes the regions of constant intensity look almost noise clean. In the case of SAR imagery, we preferred to denoise first with a Wiener filter, applying the undecimated wavelet transform next, with the Haar filter. In Figure \ref{mountain} we can see plots of 3 level Haar UDWT coefficients  of a natural image taken form the Berkeley Database, Martin et al. (2001). The undecimated wavelet transform allows to link the pixel with the coefficient, beside giving better edge localization than the DWT.

%%%%%%%%%%%%%%%%%%%%%%%%%%%%%%%%
\section{Wavelet Reduced Bayesian Net Model}
%%%%%%%%%%%%%%%%%%%%%%%%%%%%%%%%

The 2d undecimated wavelet transform consists of three bands of detail coefficients and one approximation band per level (see Figure \ref{mountain}), each of which has the same size as the original image. To statistically match the two-population property of the wavelet transform, we consider the collection of small/large wavelet coefficients from an edge detection perspective. Romberg et al. (2001), in a denoising context, modeled  the wavelet coefficients as outcomes emitted by a Gaussian mixture that approximates the marginal pdf. In our case, ``small'' and ``large'' mean differently. We continue to expect most of the nodes to be ``small'', since smooth regions produce chains of small coefficients, while singularities are chains of ``large'' coefficients. But we also expect large coefficients in close regions near the edges, so Gaussian distributions will give poor localization, making the algorithm return thick edges instead of thin ones. A more heavy-tailed distribution, like the Laplacian, will be better suited for modeling the state ``edge'', while the Gaussian will be kept as a marginal for the ``no-edge'' state.
  Thus, the conditional densities of the coefficients $w$ will be
\ben
\nonumber
	f(w|\mbox{no-edge})=\frac{1}{\sqrt{2\pi} \sigma} \exp(- \frac{w^{2}}{2 \sigma^{2}})\quad f(w|\mbox{edge})=\frac{1}{\sqrt{2}\phi} \exp(-\frac{\sqrt{2}|w|}{\phi})
\een			
 The Reduced Bayesian Net of our detector involves two independent Hidden Markov  models (HMM) for the vertical and horizontal bands, each of maximum decomposition level $T$. Thus, for each pixel $(x,y)$ we have two sequences of hidden states $S_{x,y}^{V}=\{S_{x,y,1}^{V},S_{x,y,2}^{V},\ldots,S_{x,y,T}^{V}\}$ , and  $S_{x,y}^{H}=\{S_{x,y,1}^{H},S_{x,y,2}^{H},\ldots,S_{x,y,T}^{H}\}$, $H$ and $V$ represent the horizontal and vertical bands respectively, and where each $S^{V}_{x,y,t}$  and $S^{H}_{x,y,t}$ take the ``edge'' or ``no-edge'' values.
The hidden state sequences have a 1d Markov chain property that models the Persistence property of the wavelets transforms. In this model, the observed wavelet coefficients sequence has been emitted by the hidden state sequence. The conditional distribution of the coefficients, as we said, is a Laplacian-Gaussian mixture.

The edge detector finds the two hidden state sequences for each pixel given the observed sequences of vertical and horizontal coefficients and combine them in two different ways:
\begin{enumerate}
\item The $T$ edge maps from each directional band are combined with the majority vote rule, generating  Horizontal and Vertical edge maps.
\item The Horizontal and Vertical maps are combined with the boolean ``or'' operator, see Figure \ref{flow}.
\end{enumerate}
\be
\nonumber
[\mbox{median}\{s_{x,y,1}^{V}, \ldots, s_{x,y,T}^{V}\}] \;\vee \;[\mbox{median}\{s_{x,y,1}^{H},\ldots, s_{x,y,T}^{H}\}]
\ee
Edge maps may be obtained with another choices of deterministic combinations. The Horizontal and Vertical edge maps may be obtained combining the $T$ levels of the undecimated wavelet transform with the ``and'' operator, decision that labels a pixel as edge point only if it is labeled as an edge point in all levels of the wavelet decomposition.  Applying the median operator inside each band allows us to repair sequences broke due to noise, if there is enough evidence that they are edge related. The final combination with the ``or'' operator allows us to keep independent horizontal and vertical edges, as well as corner points. The redundance of the undecimated wavelet transform keeps the information of the diagonal edges in the vertical and horizontal coefficients, suppressing the need of a third HMM model for the Diagonal band.

%%%%%%%%%%%%%%%%%%%%%%%%%%%%%%%
\subsection{Model's parameters}
%%%%%%%%%%%%%%%%%%%%%%%%%%%%%%%
Each of the HMM models has a set of parameters $\theta=\{a_{ij}, \pi_i, \phi, \sigma\}$, which has to be determined from the data. The parameters are the transition probabilities between the two hidden states $a_{ij}$, the initial probabilities in each state $\pi_i$, and the variance parameters $\phi$ and $\sigma$ of the emission densities, Laplacian and Gaussian, respectively. The means of the emission densities are set to zero, following the sparsity hypothesis of the model. We will tie the model for each pixel in the same band, considering all sequences of coefficients in each band as independent observations of the same model. Thus for the sample of sequences of random variables $W=\{W(x,y)\}$  (where $W(x,y)=\{W^{(x,y)}_1,\ldots,W^{(x,y)}_T\}$ model the noisy wavelet coefficients), our estimated parameters are the values that maximize $P(W|\theta)$:
\be
\nonumber
	\theta_{ML}= \operatorname*{arg\,max}_\theta P(W|\theta),
\ee

\subsubsection{EM algorithm for HMM parameter estimation}

We will compute the parameters using the well known Expectation Maximization paradigm, thus we will iterate until convergence the steps

\begin{itemize}
\item[] Step E: compute the conditional expectation  $E_{z|X,\theta^{(n)}}\left[\ln\left(P(X,z|\theta)\right)\right]$,
\item[] Step M: maximize the above expression on $\theta$.
\end{itemize}
\begin{thm}
The EM estimation scheme for our wavelet model consists on the following steps:  in the $(n+1)$-th step,  the forward and backward variables $\alpha_i^{(x,y)}(t)$ and $\beta_i^{(x,y)}(t)$, are computed as
\ben
\nonumber
	 \alpha_i^{(x,y)}(1)&=& \pi^{(n)}_i f(W^{(x,y)}_1|S^{(x,y)}_1=i);\;\;  \beta_i^{(x,y)}(T)= 1;\\
\nonumber
	 \alpha_i^{(x,y)}(t)&=& \left[\sum_{j}\alpha_j^{(x,y)}(t-1)a_{ji}^{(n)}\right]f(W^{(x,y)}_{t}|S^{(x,y)}_t=i),\; t=2,..,T\\
\nonumber
	  \beta_i^{(x,y)}(t)&=& \sum_{j} a^{(n)}_{ij}f(W^{(x,y)}_{t+1}|S^{(x,y)}_{t+1}=j)\beta_j^{(x,y)}(t+1),t=T-1,..,1,
\een
where $i,j=$edge, no-edge, respectively. The state probability variables, $L$ and $H$, became:
\[
	 L_i^{(x,y)}(t) = \frac{\alpha_i^{(x,y)}(t) \beta_i^{(x,y)}(t)}{P(W(x,y))}
\]
\[
H_{ij}^{(x,y)}(t) = \frac{\alpha_i^{(x,y)}(t)a^{(n)}_{ij} f^{(n)}(w_{t+1}|S^{(x,y)}_{t+1}=j) \beta_j^{(x,y)}(t+1)}{P(W(x,y))};
\]
where $P(W(x,y))=\sum_{j} \alpha_j^{(x,y)} (t) \beta_j^{(x,y)}(t)$ for any $t$.

Thus the new parameters $\theta^{(n+1)}$ are :
\ben
	\nonumber
	\pi_i^{(n+1)} &=& \frac{1}{n\, m}\sum_{x,y}L_i^{(x,y)}(1);
\een
\ben
\nonumber
	a_{ij}^{(n+1)} &=& \frac{\sum_{x,y}\sum_{t=1}^{T-1}H_{ij}^{(x,y)}(t)}{\sum_{x,y}\sum_{t=1}^{T-1}L_{i}^{(x,y)}(t)};
\een			
\ben
\nonumber
	\phi^{(n+1)} &=& \sqrt{2}\frac{\sum_{x,y}\sum_{t=1}^{T}L_{1}^{(x,y)}(t)|W_t^{(w,z)}|}{\sum_{x,y}\sum_{t=1}^{T}L_{1}^{(x,y)}(t)}\\
\nonumber
	\sigma^{(n+1)} &=& \frac{\sum_{x,y}\sum_{t=1}^{T}L_{0}^{(x,y)}(t)(W_t^{(x,y)})^2}{\sum_{x,y}\sum_{t=1}^{T}L_{0}^{(x,y)}(t)}
\een
where $n\times m$ is the size of the image.
\end{thm}

The proof of the theorem is left to Appendix A. The EM algorithm is a local descent algorithm that can be trapped in local minimum, but as we use it to construct an edge strength mask. We need to verify that the mask contains the true edges of the image, even if it provides, as all statistical edge detectors do, thick edges.
%%%%%%%%%%%%%%%%%%%%%%%%%%%%%%%%%%%%%%
\subsection{Sequence of hidden states}
%%%%%%%%%%%%%%%%%%%%%%%%%%%%%%%%%%%%%%

Given the sequence of observed coefficients $w=\{w(x,y)\}$ on a given image, and the estimated parameters $\hat{\theta}$ of the model over the image, we are interested in finding the optimal sequence of hidden states $s(x,y)=\{s^{(x,y)}_1,\ldots,s^{(x,y)}_{T}\}$ that explains $w(x,y)$, for each pixel location $(x,y)$. That is to say, we want to find the sequence $s^*(x,y)$ for which $s^*(x,y)=\operatorname*{arg\,max}_s P(s(x,y)|w(x,y),\hat{\theta})$.
\be
	s^*=\operatorname*{arg\,max}_s P(s|o,\hat{\theta})
\ee
Viterbi algorithm solves this problem, (see Appendix B). We apply this algorithm for each pixelwise sequence of coefficients, in the vertical and horizontal band, using their respective estimated parameters, obtaining $2\times T$ binary maps of edge positions, $T$ maps for each band, horizontal and vertical.

\subsection{Edge strenght Mask}
In Sun et al. (2004), strings of $T$ labeled coefficients produce a label ``edge'' only if they have all the evidence in favor of being an edge, that is, the string produced by a branch in the wavelet tree is an unbroken sequence of $E$ states.  In our case,
 we compute the most likely state maps that may have emitted the observed image, and decide that a pixel is a candidate to edge point if it is flagged as one  in more than one half of the state maps of any of the directional bands considered.

 The final edge strength map has a constant rate of false alarms but a low misclassification rate, so the second stage in edge detection, that is made usually with active contours, can successfully locate the real edges.

%%%%%%%%%%%%%%%%%%
\section{Experimental Results}
%%%%%%%%%%%%%%%%%%

The appropriate noise model for speckled images is the multiplicative one. We apply the logarithmic transformation to obtain an image with the appropriate additive model, in order to consider Laplacian Gaussian mixtures as the correct emission distributions. Natural optical images with high variance Gaussian noise added are examples that fit the additive noise model and have edges on all scales, providing good challenge for our algorithm. Also, there are several classical databases of images for edge detection which provide benchmark images with ground truth. In Figure \ref{medidas} we show three different Ground Truth images, made by different people, that show the different scales that edges of natural images have. The database was made adding seven levels of Gaussian noise, $\sigma=0,5,10,20,30,40,50$, to the test images, downloaded form the Berkeley Database \cite{Martin}.

 We will compare our edge detector, WBND, with a simple but effective wavelet based thresholding scheme (HTHW), also based on the undecimated Haar wavelet transform, and Canny edge detector, the usual choice among practitioners. The deterministic algorithm HTHW was included to show the implicit power for edge detection that the projection in wavelet space possesses, and to show the reducing performance effect that noise produces on its detection capability.

In a second experiment, we will work with several SAR images obtained from different sensors, showing different kinds of textures at different resolutions. We will work with an ESAR PRI (high resolution SAR image) of ice land, produced to study seasonal polynyas, regions of open water inside the Artic casquet, an ESAR VV three looks (moderate resolution SAR image) of a region outside the city of Munich, produced to study automatic generation of road maps, and finally, an ESAR full polarimetric complex image without any processing, with several kinds of textures, from a country-side region of Germany, which has been extensively studied in other recent edge detection proposals, Giron et al. (2012).

We will begin detailing the HTHW algorithm we use for making the comparison's simulation.

\subsection{Hard Thresholding Haar wavelet detector (HTHW)}
%\begin{figure}[h!]
%\includegraphics[width=15cm]{Paper_StatCOMP_fig4.pdf}
%\caption{Flowchart of a simple wavelet based edge detector.}
%\label{HTHW}
%\end{figure}
%%%%%%%%%%%%%%%%%%%%%%%%%%%%%%%%%%%%%
Many well-known edge detectors, like Prewit, Roberts, Sobel and Canny, are based on the principle of matching local image regions with specific edge patterns. These edge patterns are in fact impulse responses of high-pass filters that suppress homogeneous and smooth regions while emphasizing discontinuities. The edge detection is performed by convolving the image with a set of directional patterns, followed by thresholding, Kitanovski et al. (2009). Wavelet based edge detectors also involve these ideas.

 We implemented a simple edge detector thresholding the undecimated Haar wavelet decomposition at the first level, which is roughly like computing the first derivative in three directions, horizontal, vertical and diagonal. We label a pixel location as edge if the maximum of the sizes of its related three coefficients surpass a given threshold. We also keep the information about the orientation of the biggest coefficient to apply non-maximal suppression as Canny's method does.% In Tello Alonso et al. (2011), a similar scheme is proposed in several levels of the stationary Haar wavelet transform, combining them with a pointwise product, to ``enhance" the edges of SAR images. The pointwise product between levels is equivalent to keep only edge points labeled as ``L'' in all levels. In Figure \ref{HTHW} we can see a flowchart of the HTHW edge detector.

 Our statistical algorithm, WBND, select the coefficients as being in the state ``edge'' or ``no-edge'' not only when their amplitude is large (as HTHW does), but also when it is moderately large but still has high probability of being produced by the ``edge'' emission distribution.
% Besides, we consider more than one level in the wavelet transform, to take advantage of the persistence property of the coefficients.

\subsection{Initial value selection}

The thresholds for Canny and  HTHW method are image dependent, and chosen using the Pratt-quality curves method discussed on Fernandez-Garcia et al. (2004). We moved the parameters in a one dimensional fashion, 100 steps, and let the quality measures choose the best map. In the case of Canny, we fixed the noise smoothing parameter to the default, and moved only the threshold, using the Matlab function ``edge''. In the case of HTHW, there is only one parameter to move, since it does not have a  built-in smoothing procedure. Nevertheless, we applied a median filter to the image before computing the edge map, in order to compete in a fair way with Canny.

For our method, WBND, we choose non-informative priors for the state values, and transition probabilities that strongly penalize the change from one state to the other.
\be
\nonumber
	\pi^{(0)}=[0.5\;\; 0.5]', \qquad A^{(0)}=\displaystyle \left[ 0.95 , 0.05; 0.2 , 0.8\right].
\ee
The initial variance parameters of the conditional emission distributions, $\phi^{(0)}$ and $\sigma^{(0)}$, were carefully chosen from histograms of the coefficients of each image.

\subsection{Quality measures}

In this section we define three quality measures well known in the computer vision literature. These measures need ground truth maps to produce the comparison output,  which are always available when working with artificial images.
\begin{figure}[h!]
\includegraphics[width=15cm]{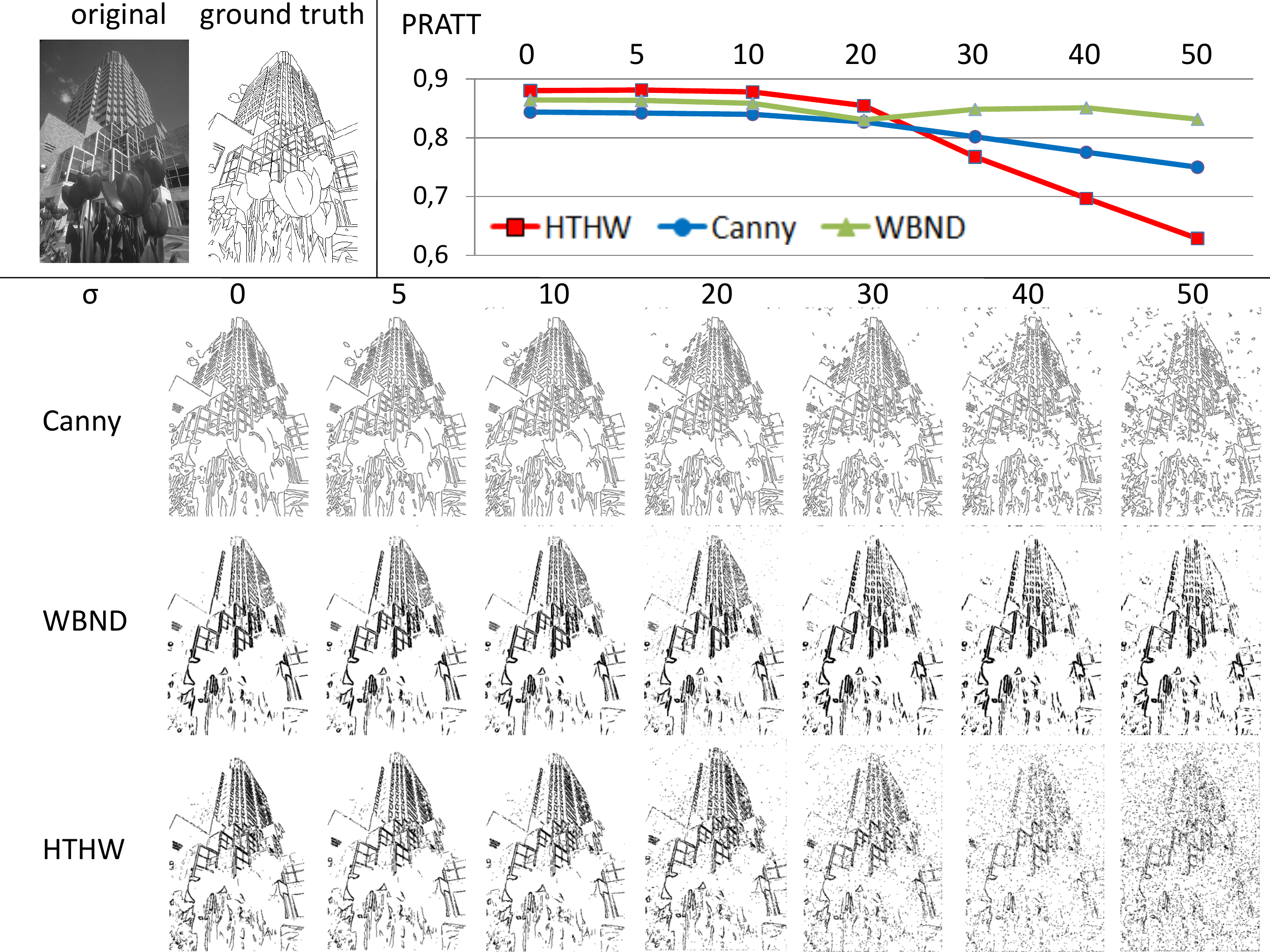}
\caption{Edge maps of Building image with 7 levels of noise added: $\sigma=0,5,10,20,30,40,50$, made with Canny, WBND and HTHW methods. On top, Pratt's quality curve over noisy images, for all three methods. The edge maps of Canny and HTHW methods were selected by Pratt measure, as the peak of the curve made moving the threshold parameter 100 times. The edge maps of WBND method was the automatic output of the algorithm for each noisy input image.  }
\label{edificio1}
\end{figure}
\subsubsection{Pratt's quality measure}
Let us consider $V$ the ground truth of an image $I$, and $B$ another computed edge map over the same image, $E_V$ and $E_B$ the set of edge points of both maps and $\rho$ be the Euclidean distance. Thus, Pratt's measure is defined by
$$P_V(B)=\displaystyle\frac{1}{\max\{\#E_V,\#E_B\}}\displaystyle\sum_{s\in E_B}\displaystyle\frac{1}{1+\alpha \ \rho(s,E_V)^2}$$
with $\alpha=\frac{1}{9}$.
\subsubsection{Baddeley's error measure}
  Baddeley's distance is defined by
$$\Delta_{\omega}^p(V,B)=\left[\displaystyle\frac{1}{nm}\displaystyle\sum_{s\in E_V}|\omega(\rho(s,V))-\omega(\rho(s,B))|^2\right]^{1/2}.$$
with $\omega(t)=\min\{t,5\}$.
\subsubsection{Kappa's quality measure}The index Kappa is defined by
$$\kappa=\displaystyle\frac{P_o-P_e}{1-P_e}$$
where $P_0$ is the observed proportion of edges and $P_e$ is the expected proportion of edges in the image, one computed over the estimated edge map $B$ and the other computed over the ground truth map $V$.

\subsection{Results on optical images}

\begin{figure}[h!]
\includegraphics[width=15cm]{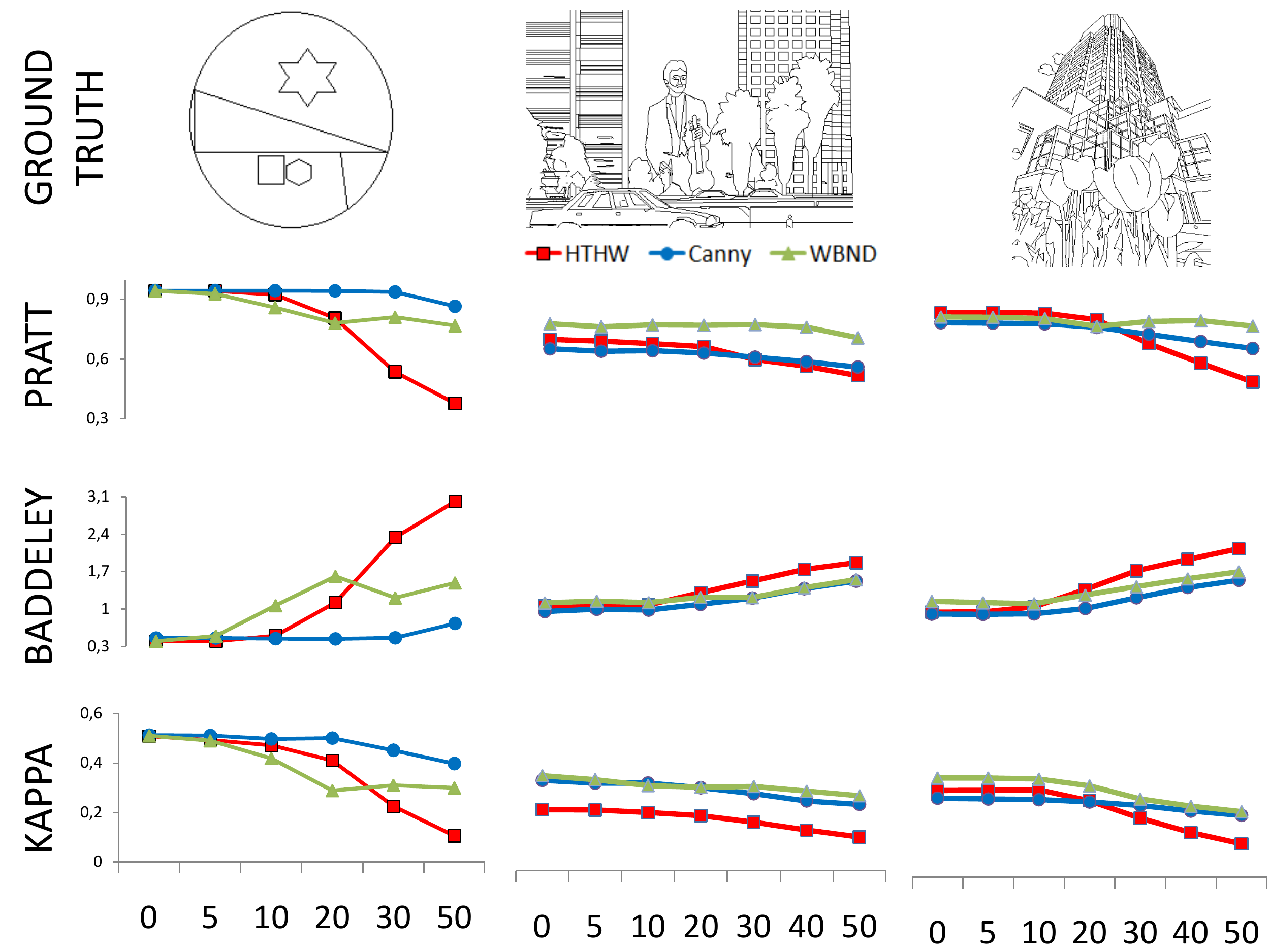}
\caption{Quality curves of  Canny, WBND and HTHW methods, computed over edge maps of Star, Building and Giant image with 7 levels of noise added: $\sigma=0,5,10,20,30,40,50$. The measures are Pratt's quality measure, Baddeley's error measure and Cohen's Kappa quality measure. All these measures need knowledge of Ground Truth to deliver an output.}
\label{medidas}
\end{figure}

We compute 50 edge maps per image sampling the parameter space of Canny and HTWD. We add 7 levels of noise to each of these images, and compute the distance from them to the Ground Truth, generating a 100 point quality curve for each image. The edge map that makes the maximum value over such curve is the map chose to compete with WBND. In Figure \ref{medidas} and \ref{edificio1} we plot the maximum value of the measure (over the 100 edge maps) as a function of the noise level.

In Figure \ref{medidas} we can see plots of the three quality measures as functions of seven levels of noise, computed over three images: a simple synthetic image, Star, and two natural images, Giant and Building,  with detailed hand made ground truth. The images has very different scaling behavior in their structural edge system. Star, the first panel of Figure \ref{medidas}, is a very simple artificial image, with very thin edges and no texture. The other two images were natural images that depict man made constructions, with edges in several resolutions.

The positive quality measures, Pratt and Kappa, score  maximum quality with the 1 value. Baddeley is an error measure, scoring maximum quality with zero. Keeping in mind these ideas,  the measure curves have different interpretations from image to image. In general, with low level of noise, the optimized Canny found the boundaries of the objects almost perfectly, but with high level of noise, errors add up and quality worsens. The edges outside the boundary, as well as the texture, are not well resolved by Canny's method in any image.

This  edge strength maps are the first step in boundary and structure detection. The next step would be to apply an active contour method to deliver a one pixel wide, continuous edge structure. In Figure \ref{edificio1} we can see that WBND gives a strong set of lines to follow. The other two methods give too many false alarms outside the boundary of the object. We should stress again that the maps shown are the best maps chosen by the quality measures contrasting them against the Ground Truth.

\begin{table}[h!]
\centering
\begin{tabular}{|c|c|c|c|c|c|c|c|c|c|c|}
  \hline
  & &\multicolumn{3}{|c|}{Pratt}&\multicolumn{3}{|c|}{Baddeley} &\multicolumn{3}{|c|}{Kappa} \\ \hline
  % after \\: \hline or \cline{col1-col2} \cline{col3-col4} ...
BD Id&Image	  &HTHW	& Canny 	&	WBND &HTHW	& Canny 	&	WBND &HTHW	& Canny 	&	WBND     \\ \hline
86000& building	&  	0,427	 &  0,529	&	0,629  &1,908	&1,734		&1,757   &0,098  &0,173	&0,222  \\
119082&giant	    &   0,563    &	0,537	&	0,649  &1,846	&1,721	    &2,257   &0,190	 &0,269	&0,230\\
 69015&koala      &	0,454	 &	0,494	&	0,606  &2,298   &2,177		&1,985   &0,077	 &0,172	&0,281\\
187071& mountain   &	0,391	 &	0,488	&	0,655  &1,740	&1,911	    &1,459   &0,109	 &0,204	&0,307 \\
24077& monk       &	0,508	 &	0,473	&	0,558  &2,031	&2,153	    &2,605   &0,133	 &0,139	&0,177 \\
42049&bird        &	0,517	 &	0,473	&	0,928  &1,235	&1,182	    &0,795   &0,342	 &0,391	&0,606 \\
106024& penguin    &	0,262	 &	0,327	&	0,476  &1,584	&1,682	    &2,051   &0,094	 &0,155	&0,229 \\
 300091& surf      &	0,311	 &	0,436	&	0,745  &1,141	&1,347	    &0,921   &0,153	 &0,227	&0,423 \\
108082&tiger       &	0,156	 &	0,193	&	0,208  &1,493	&2,261	    &1,916   &0,017	 &0,046	&0,064 \\
  \hline
\end{tabular}
\label{ta}
\vspace{.1in}
\caption{Maximum quality measurements of edge maps made with images from the Berkeley segmentation database, obtained by moving the parameters in a wide range as explained in Fernandez-Garcia et al. (2004).}
\end{table}

Table 1 shows values of the three measures computed over edge maps of 9 images selected from the Berkeley database with the maximum level of noise added, to simulate signal to noise ratios similar to a SAR image. Quality values are computed referring to the most detailed segmentation that the Berkeley database can provide for the selected image, which is usually the boundary of the main objects, with few details. We selected pictures of animals with different textures, fur and feathers, pictures of landscapes and sea waves, and pictures of man made constructions, as buildings, statues and cars. WBND consistently shows better values than the other two methods.

\section{Sar image data}

\subsection{ERS SAR PRI image data of Artic Ice}
In this section, we will apply our edge detector to a portion of a high resolution ERS SAR PRI image (25 m spatial resolution), VV polarization,  taken on February 15, 1994, on the southern part of the coastal Artic, the Franz Josef Land. This image of ice land includes a polynya visible diagonally from the top left to the bottom right of the image, see Dokken et al.(2002) for details of the imagery and this important environmental problem.

A polynya is a geographical term for an area of open water surrounded by sea ice, i.e. unfrozen sea within the ice pack. Knowledge of the distribution and frequency of coastal polynyas and leads is important in understanding large-scale climate processes in the Arctic Basin. Traditionally, satellite passive microwave sensors are used for polynyas studies due to their temporal and spatial coverage. Due to the typical size of polynya features (on average 6 km in width) a relatively high spatial resolution of the satellite sensor is required. The active synthetic aperture radar (SAR) enables high resolution imaging (25 m spatial resolution) of the geographic region of interest during night and day and through the often-occurring cloud cover above these features. In Dokken et al. (2002), several algorithms for segmentation of Artic ice images were compared with a new segmenter, called the SAR polynya algorithm,
to delineate open water, new ice, and young ice, and to define the size and shape of polynyas. This algorithm uses wavelet projection in two different ways, to recover texture signatures and to extract edges. Such a complex type of algorithm is very common in SAR processing studies, since region detection in the images is very difficult.
\begin{figure}[h!]
\includegraphics[width=15cm]{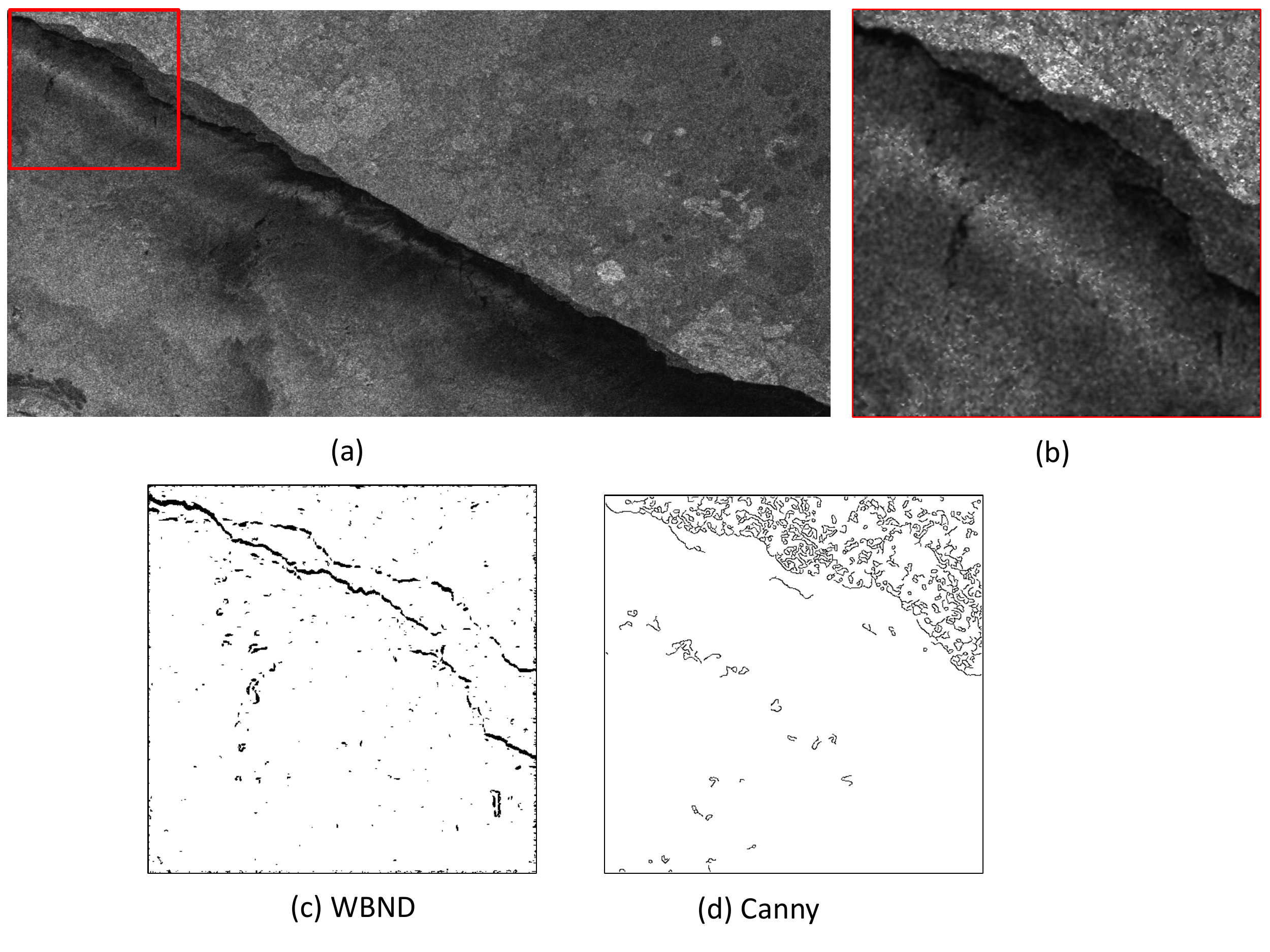}
\caption{First row, panel (a),  an ERS SAR PRI image taken on 15 February 1994 of the southern part of the Franz Josef Land including a polynya diagonally visible from the top left to the bottom right of the image. Panel (b) on the right, a small region of the top left part of the image. Second row, Panel (c)   WBND edge map of the region selected, (d) Canny's edge map of the same selected region, with optimized parameter to reduce noise related artifacts. WBND final parameters have been selected automatically by the EM algorithm, initial parameters set with histogram quantiles.}
\label{ice}
\end{figure}

Dokken et al. (2002) suggest that enhancements of the SAR polynya algorithm could be  to combine different image representations (e.g. curvelets and ridgelets) in order to improve the delineation of young ice types inside the polynya region. We may suggest  replacing the last loop of the algorithm, the deterministic wavelet based edge detection, by our Hidden Markov Model approach based on our simulations, where we clearly outperform wavelet thresholding techniques using statistical models. In Figure \ref{ice} we can see one of the images studied in Dokken et al. (2002), with a the polynya in it. In that image we selected a small region and computed Canny's edge map and WBND edge map.  We can report that our algorithm delineate the region of thin ice much better than Canny's algorithm, allowing a segmentation of the region where the polynia should lay.

\subsection{ERS SAR image data from city}

\begin{figure}[h!]
\includegraphics[width=15cm]{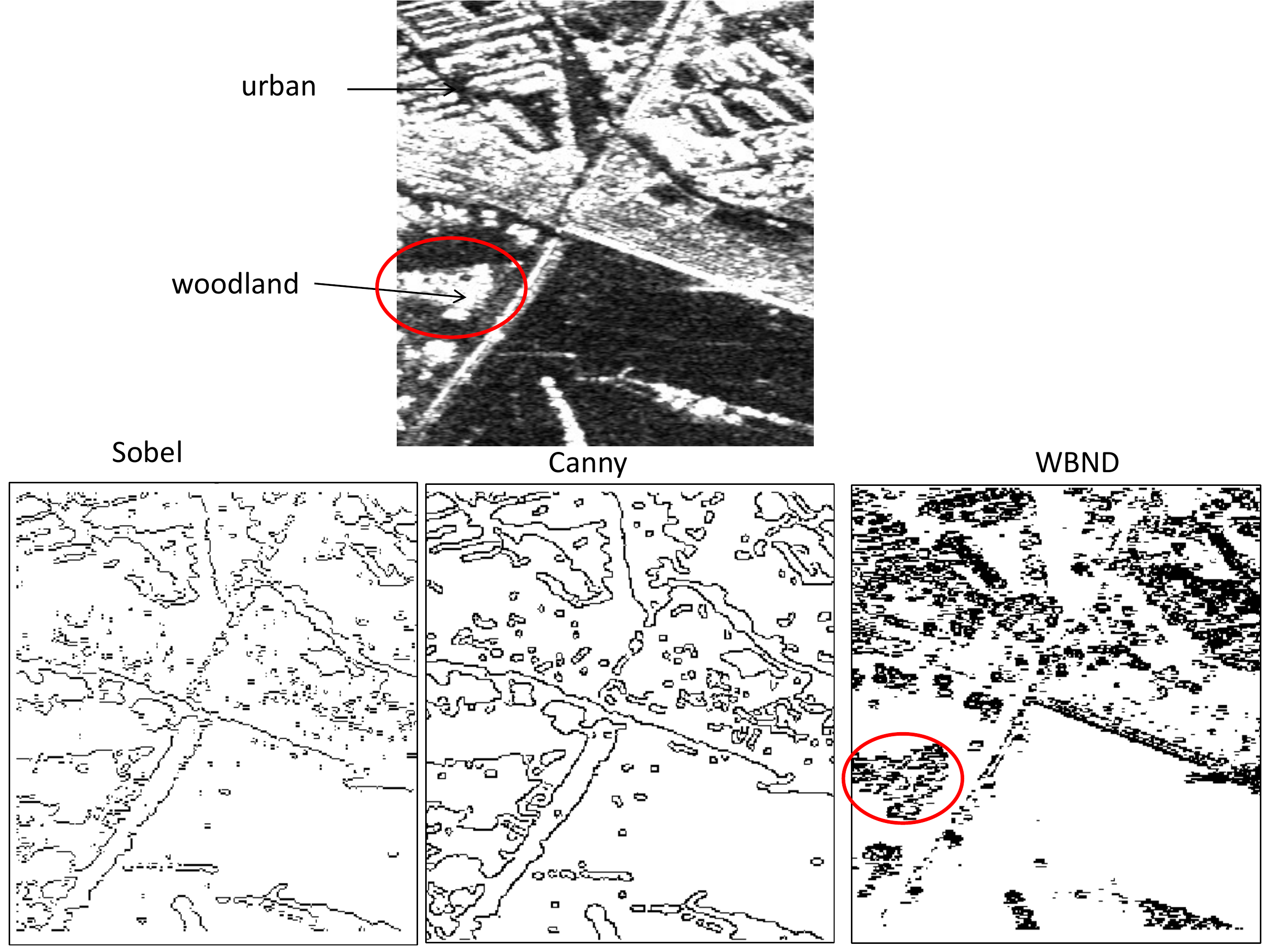}
\caption{In the first row, HH band of an approximated 3 looks real ERS SAR image showing an urban area from
the city of Munich. In the second row, Sobel, Canny and WBND edge maps. We have remarked a woodland region that were well segmented by our algorithm, but that were not isolated by the other two detectors.}
\label{munich}
\end{figure}

In this section we apply our algorithm to an ERS SAR image containing urban and woodland areas.
We selected a $373\times 722$ subimage, taken from a 1500 x 4000 ERS-1, L
band, HH polarization image with an estimated number of
looks equal to 2.95, corresponding to Munich, Germany. In Figure \ref{munich} we show the image, and the resulting edge map estimating only one model for the whole piece. The image shows several highways and roads, which are reasonably well detected by the algorithm. Pixel resolution is low, around 100 square meters, since it was averaged three times to reduce speckle, so an accurate road map could only be made postprocessing the edge map with subpixel accuracy. Canny's and Sobel edge maps have thinner lines than the WBND edge map, but do not isolate the regions accurately. In Figure \ref{munich} we can see a small portion of woodland which was only well detected by our proposed algorithm.

\subsection{Full Polarimetric Complex ESAR imagery }

In the previous examples we worked with images with only one polarization band available. In many cases, combination of polarization signatures is important to reveal differences in texture, Mortin et al. (2012).
\begin{figure}[h!]
\includegraphics[width=15cm]{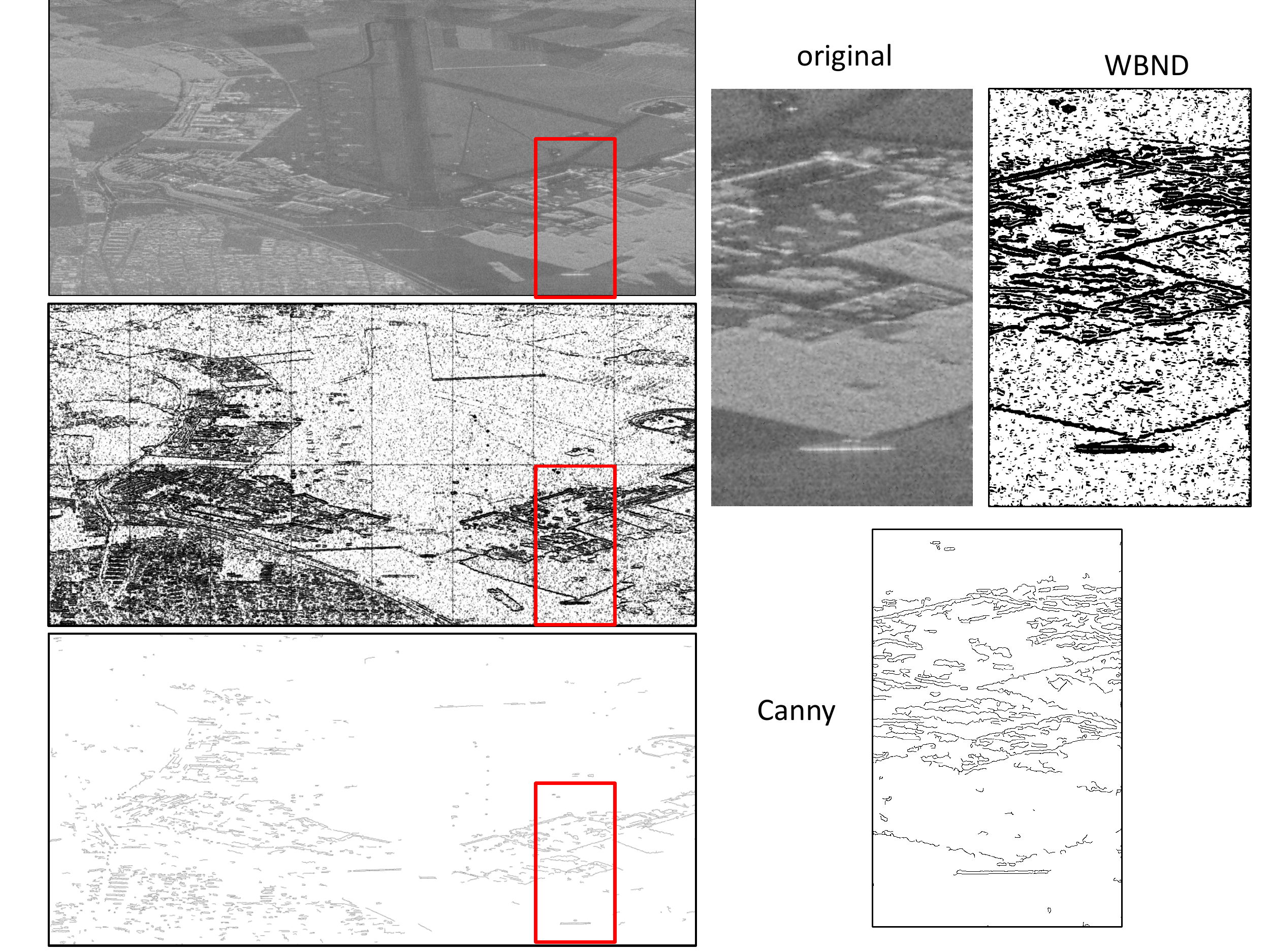}
\caption{ First column, first panel,  log-amplitud from the HH polarization band of a  E frequency band SAR image from a zone near the city of Webling, Bayern, Germany, $4000\times1499$ pixels, with an urban region selected. Second and third panels:  WBND map of the whole image, computed in parallel fashion with the same initial parameters, and  Canny edge map computed with optimized parameters.  Second column: enlarged image of selected region and WBND and Canny's edge maps. We should notice that our proposal generate closed thick edges around regions, while Canny's fail to close region boundaries.  }
\label{sar1}
\end{figure}

In this section we will work on a full polarimetric E-SAR image that shows three different textures, agricultural land, cities and woodland. The image considered is a raw $\mathcal C^4$ complex, E band SAR image from a zone near the city of Webling, Bayern, Germany. The image has $ 1499\times4000$ pixels, and shows a variety of textures and man-made constructions, as we can see in Figure \ref{sar1}. We divided the image in 16 pieces, in order to compute the map in a parallel implementation with the same initial parameters.  In Figure \ref{sar1} we can see the whole map computed over the HH polarization band of the image, and as we found out in the previous subsection, the city roads and woodland regions, corresponding to heterogeneous and extremely heterogenous textures, were well detected by the algorithm. Canny's algorithm produces thinner edges than our's, but misses a large part of the boundary of heterogeneous regions.

\begin{figure}[h!]
\includegraphics[width=15cm]{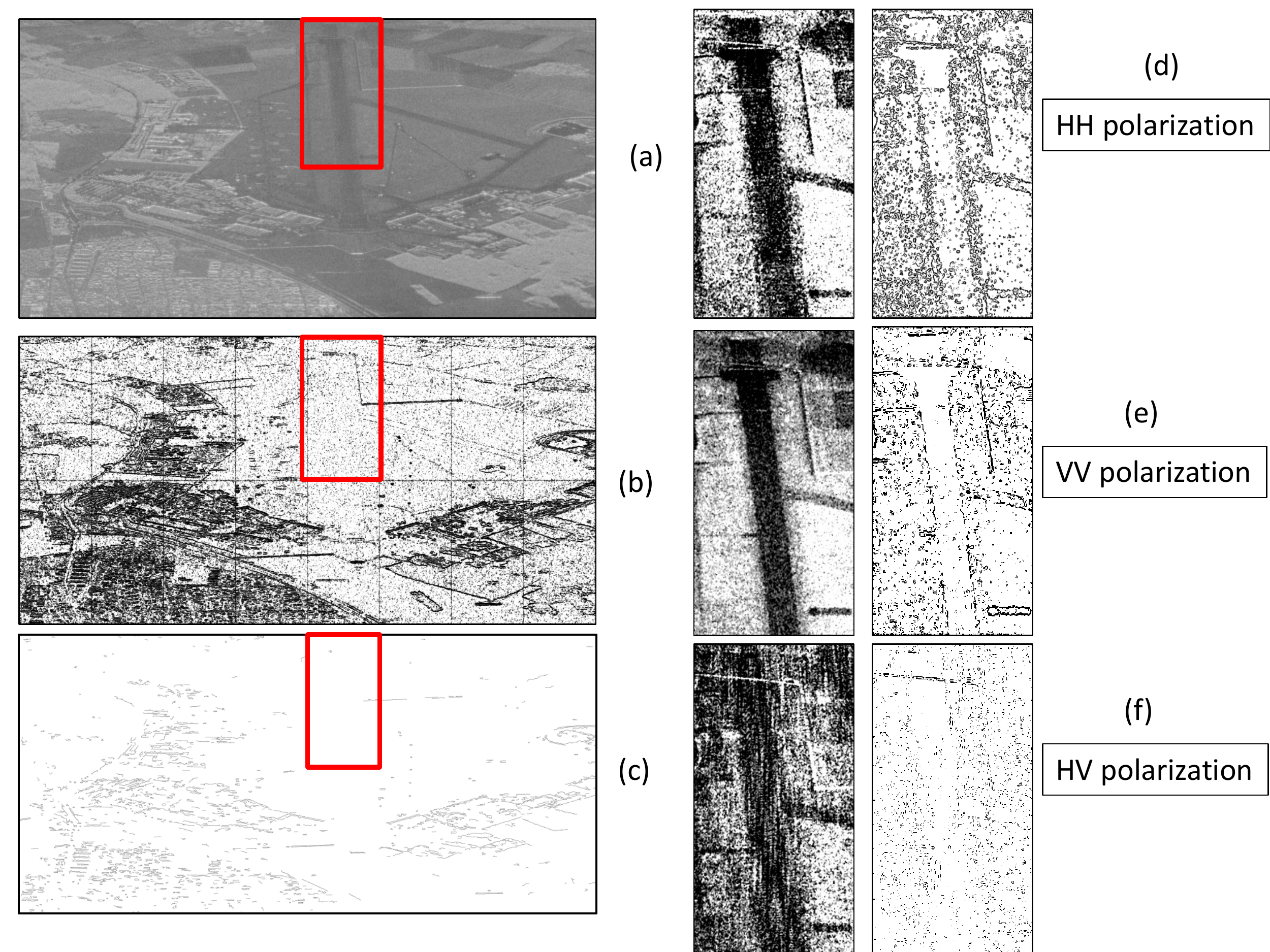}
\caption{First column, first panel, log-amplitud from the HH polarization band of a  E frequency band SAR image from a zone near the city of Webling, Bayern, Germany, $4000\times1499$ pixels, with a region of a low contrast runway selected. First column, second and third panel: WBND map of the whole image, computed in parallel fashion with the same initial parameters, and Canny's edge map made with optimized parameters.  Second column: HH-band, VV-band and VH band region of the runway, after histogram equalization,  and related WBND edge map, in that order.}
\label{sar2}
\end{figure}

In Figure \ref{sar2} we can see another region selected in the original image, a piece of a long runway, and the same region remarked on Canny's and WBND edge maps. Non of the algorithms could detect that long runway.  An explanation for this is that the logarithmic transformation considerably reduces the natural contrast of the HH image. In the two other examples we have shown, the images had a better resolution, so the logarithmic transformation did not diminish the detection power of the algorithm. This image has three other polarizations available, but it has not been processed to reduce speckle, has no number of looks,  so it is the noisiest product available.

We selected the region of the runway we showed in Figure\ref{sar2}, in all polarizations available, and re-run the algorithm. In the second column of the Figure \ref{sar2}, we can see the runway found, in all images. Nevertheless, the runway is much better defined in the edge map corresponding to VV polarization. Nevertheless, keeping edges from all the different maps made with different polarizations images increases noise in the final map, so combination should be made with a penalization. Man-made constructions were well detected  with HH polarization images, low contrast regions (or natural homogeneous areas as the polynyas), were better detected fine tuning VV polarization regions, and VH have not well defined edges.

The proposal of a full polarimetric edge detector is beyond the scope of this paper, but the use of a multidimensional complex wavelet transform could be the way to avoid the logarithmic transformation, maintaining the contrast range of the image in all polarizations. To our best knowledge, no statistical model has been fitted to the coefficients of a multidimensional complex wavelet transform, relating edge and no-edge states, and we are currently studying different transformations and algorithmic implementations, searching for the best suited for edge detection.

%%%%%%%%%%%%%%%%%%
\section{Conclusion and Discussion}
%%%%%%%%%%%%%%%%%%

In this work we described a new approach to edge map construction in SAR images using hidden Markov Models on the undecimated wavelet domain. Most Hidden Markov models on the wavelet domain have been defined for denoising applications, and they traditionally favor orthogonal transformations. For edge detection, shift invariant transforms are better suited for the task than orthogonal ones, since redundance allows better edge localization and tracking along the wavelet representation. Also, modeling coefficients with a Laplacian-Gaussian mixture implies that an implicit denoising is taking place at the same time edge related coefficients are selected. This explains the reason why our algorithm maintains a good performance in our simulations when  Gaussian noise with high variance was added.

Our algorithm  is based on modeling the coefficients of three consecutive levels of the horizontal and vertical detail bands of the undecimated wavelet transform with a three step Markov chain with two possible hidden states, ``edge'' and ``no-edge'', which jointly fit a Laplacian-Gaussian mixture with zero mean. This two independent models conformed a  reduced Bayesian Net on the undecimated wavelet domain, which is simple but effective in capturing the two main properties of edge related wavelet coefficients, persistence and non gaussianity. The final edge map is produced combining the most likely sequence of states giving the estimated models, keeping all vertical, horizontal and corner edges.

We investigated the performance of a deterministic wavelet thresholding detector, and the well-known Canny edge detector along with our WBND algorithm for edge detection in both synthetic images and natural images with available ground truth. From the experimental results on the artificial image, it was observed
that Canny detector performed slightly better than WBND and HTHW when moderated Gaussian noise was added, and that our method had an increasingly better performance when higher levels of noise were added. The performance was measured using Pratt's, Baddeley and Kappa classical quality measures. On natural images with considerably more edges at different resolution sizes than synthetic images, our WBND detector appeared to be the most robust to variations in noise,
performing well in all noise configurations tested.

We also report resulting edge maps on  three different resolution and polarization SAR images:  a
VV polarization high resolution ERS SAR PRI image of iced land, an HH polarization L frequency ERS image with three looks of a region outside the city of Munich, and  a low resolution SRC raw complex image, the noisiest kind of product, with the HH, VV and VH bands available. Edge detection was successful in all extremely heterogeneous textured regions, but the logarithmic transformation flatten the contrast on the SRC image, introducing detection errors in the homogeneous regions. Increasing contrast on homogeneous regions, and considering all polarizations, we can report that  edge maps were better formed on the VV band than on the HH band, and combinations of the two edge maps resulted in very noisy outputs.
%We thus suggest to use HH band polarization for regions of heterogeneous texture, while using VV polarization for homogeneous regions, if all bands are available.

The proposal of a full polarimetric edge detector is beyond the scope of this paper, but the use of the undecimated or stationary complex wavelet transform could be the way to avoid the logarithmic transformation, maintaining the contrast range of SRC images.  We are aware of Choi et al. (2002) work on Hidden Markov Trees model for Kingsbury's dual complex wavelet transform, but to our best knowledge, no statistical model has been fitted to the coefficients of the complex wavelet transform of a multidimensional complex image, relating edge and no-edge states, and we are currently studying different transformations and algorithmic implementations, searching for the best suited transform for edge detection.

\end{document}